\documentclass[a4paper]{article}

\usepackage[english]{babel}
\usepackage[utf8x]{inputenc}
\usepackage[T1]{fontenc}
\usepackage{float}
\usepackage{caption}
\usepackage{subcaption}
\usepackage[a4paper,top=3cm,bottom=2cm,left=3cm,right=3cm,marginparwidth=1.75cm]{geometry}
\usepackage[ruled,vlined]{algorithm2e}

\usepackage{amsmath}
\usepackage{graphicx}
\usepackage{url}
\usepackage[colorinlistoftodos]{todonotes}
\usepackage[colorlinks=true, allcolors=blue]{hyperref}

\newcommand{\pkg}[1]{{\normalfont\fontseries{b}\selectfont #1}}
\let\proglang=\textsf

\numberwithin{equation}{subsection}

\title{\textbf{Categorical exploratory data analysis on goodness-of-fit issues.}}
\author{\textbf{Sabrina Enriquez}, \textbf{Fushing Hsieh}\\ \url{seenriquez@ucdavis.edu}, \url{hsieh@ucdavis.edu}}

\begin{document}
\maketitle
\begin{abstract}
If the aphorism: "All models are wrong"- George Box, continues to be true in data analysis, particularly when analyzing real-world data, then we should annotate this wisdom with visible and explainable data-driven patterns. Such annotations can critically shed invaluable light on validity as well as limitations of statistical modeling as a data analysis approach. In an effort to avoid holding our real data to potentially unattainable or even unrealistic theoretical structures, we propose to utilize the data analysis paradigm called Categorical Exploratory Data Analysis (CEDA). We illustrate the merits of this proposal with two real-world data sets from the perspective of goodness-of-fit. In both data sets, the Normal distribution’s bell shape seemingly fits rather well by first glance. We apply CEDA to bring out where and how each data fits or deviates from the model shape via several important distributional aspects. We also demonstrate that CEDA affords a version of tree-based p-value, and compare it with p-values based on traditional statistical approaches. Along our data analysis, we invest computational efforts in making graphic display to illuminate the advantages of using CEDA as one primary way of data analysis in Data Science education.
\end{abstract}

\section{Introduction}
Goodness-of-fit testing is one essential topic of data analysis and mathematical statistics. In data analysis, we want to know that, if an existing distribution function is chosen as one perspective for reading a data set in hand, we would like to discover where and how data’s empirical distribution is explicitly fitting or deviating from this known distribution functional form. In mathematical statistics, such hypothesis testing is the bread and butter of statistical education. This testing involves many key concepts and methodologies: the likelihood principle, testing statistic constructions, null distribution, p-value and several others.  This topic indeed has been routinely practiced in data analysis and heavily developed in statistical theory since the beginning of 20th century \cite{Pearson}. After more than 120 years into its history, can we now add new perspectives and understanding into this topic? In this paper, we demonstrate the advantages of adding new machine-learning based perspectives to evaluating goodness-of-fit.

A known distribution function $F(x)$ on $R^1$, such as Normal distribution $N(\mu, \sigma^2)$, is indexed by only a few parameters, such as mean $\mu$ and variance $\sigma^2$. These real valued parameters are usually not the chief concern in goodness-of-fit testing. The chief concern of this hypothesis testing is placed on its functional form of $F(x)$. For instance, in the Normal distribution example, its functional form $F(x)$ is explicitly displayed via:
\[
F(x)=\Phi(x|\mu, \sigma)=\frac{1}{\sqrt{2\pi \sigma^2}} \int^x_{-\infty}\exp{\frac{-(z-\mu)^2}{2\sigma^2}} dz ; x\in R^1.
\]
This functional form of $\Phi(x|\mu, \sigma)$ on $R^1$ is conventionally taken as an infinite dimensional parameter. 

In contrast, this conventional view of a function is somehow applied to an empirical distribution function $\hat{F}_n(x)$, which is $n$-dimensional with $n$ being equal to the sample size. Here $\hat{F}_n(x)$ is built based on a data set $\{x_1, x_2,…, x_n\}$ in the following manner: with indicator function $1_{[x_i \leq x]}$ for all $i=1,..,n$,
\[
\hat{F}_n(x)=\frac{1}{n}\sum^n_{i=1} 1_{[x_i \leq x]}.
\]
From a data structure perspective, with varying $n$, $\hat{F}_n(x)$ is of unstructured data type in the sense that it can’t be expressed by a fixed vector (independent of $n$). 

Thus, the fundamental setting underlying any goodness-of-fit testing is: evaluating the degree of matching between an infinite dimensional parameter $F(x)$ with an unstructured data $\hat{F}_n(x)$ for all $x\in R^1$. This is a rather complex problem from data structural perspective. 

Can a single one-dimensional summarizing statistic, as commonly developed in statistics literature, fully capture the complexity of goodness-of-fit-testing? A summarizing statistic typically collects one common kind of departure of $\hat{F}_n(x)$ from $F(x)$. So, a testing statistic is designed and proposed to effectively deal with one specific kind of departure. But it becomes rather ineffective when facing other kinds of departures, which are perpendicular to the designed one. 

In reality, there might be multiple kinds of departures occurring at the same time when comparing $\hat{F}_n(x)$ with $F(x)$.  In such a case, it becomes natural to ask simple questions like: What kinds of remedial approaches can machine learning (ML) inspire us upon this essential task in real-world data analysis? What if we construct a goodness-of-fit testing platform, instead of a testing statistic? Since we not only want to evaluate the degrees of departure of $\hat{F}_n(x)$ from $F(x)$, but we also want to see and explain where and how the departures occur, a testing platform is more informative. That is, this platform ideally will enable us to quantitatively evaluate one single overall degree of departure. This quantitative degree will realistically serve the function of statistical p-value, but without being subject to unrealistic assumptions or structures, such as being assumed independently identically distributed (i.i.d), which the statistical p-value is most often subject to. 

In this paper, we endeavor to address aforementioned natural questions. We develop a matrix-platform for fully exhibiting identified characteristics of $\hat{F}_n(x)$ for the purpose of goodness-of-fit testing, and create a new tree-based algorithm for computing an ML-p-value. We illustrate our developments through two examples of real-world data sets.

\section{Traditional methods of data fitting}
We consider a couple of well known \cite{trad} traditional methods for evaluating normality of experimental data. First we will evaluate using Q-Q plots as a graphical technique, and then we will utilize tests that can yield p-values: the Shapiro-Wilk normality test, Pearson's chi-squared test, and the Kolmogorov- Smirnov test. In this section we will point out chief features of these methods and simultaneously make clear what capacities these traditional methods are lacking when facing real-life data scenarios. Here these capacities are all related to categorical patterns underlying the departure between observed data and the modeling distributional structures of choice.

Patterns of categorical nature include asymmetry, gap, jump, hump and other non-smooth high frequency structures. Such kinds of categorical pattern are hardly concerned and accommodated in classic summarizing statistics due to smallness of data sets. It is understandable when number of data points is small, most of data analysis is conducted by embracing the approximating sense. In sharp contrast, when sample size is large or even huge, this approximating sense of data analysis might be no longer valid. Since all goodness-of-fit testings are more or less bound to be rejected due to all sorts of existential categorical patterns. If these categorical patterns are remained hidden throughout the course of data analysis, and all these patterns are indeed harmless to the designated goal that we want to achieve, then we miss effective summarizing statistics via a modeling distributional structure.

Therefore, at this Big-Data era, the key task of goodness-of-fit testing is not as much aiming at determining whether a modeling distributional structure being acceptable or not as discovering all potential categorical patterns underlying the departure between observed data and the modeling distributional structures. By discovering these potential categorical patterns, we then figure out which patterns are detrimental, and which patterns are harmless, to our designated goal of data analysis. This is the brand new perspective of goodness-of-fit testing.

\subsection{Q-Q Plot}
\begin{figure}[H]
    \centering
    \includegraphics[width=.9\textwidth]{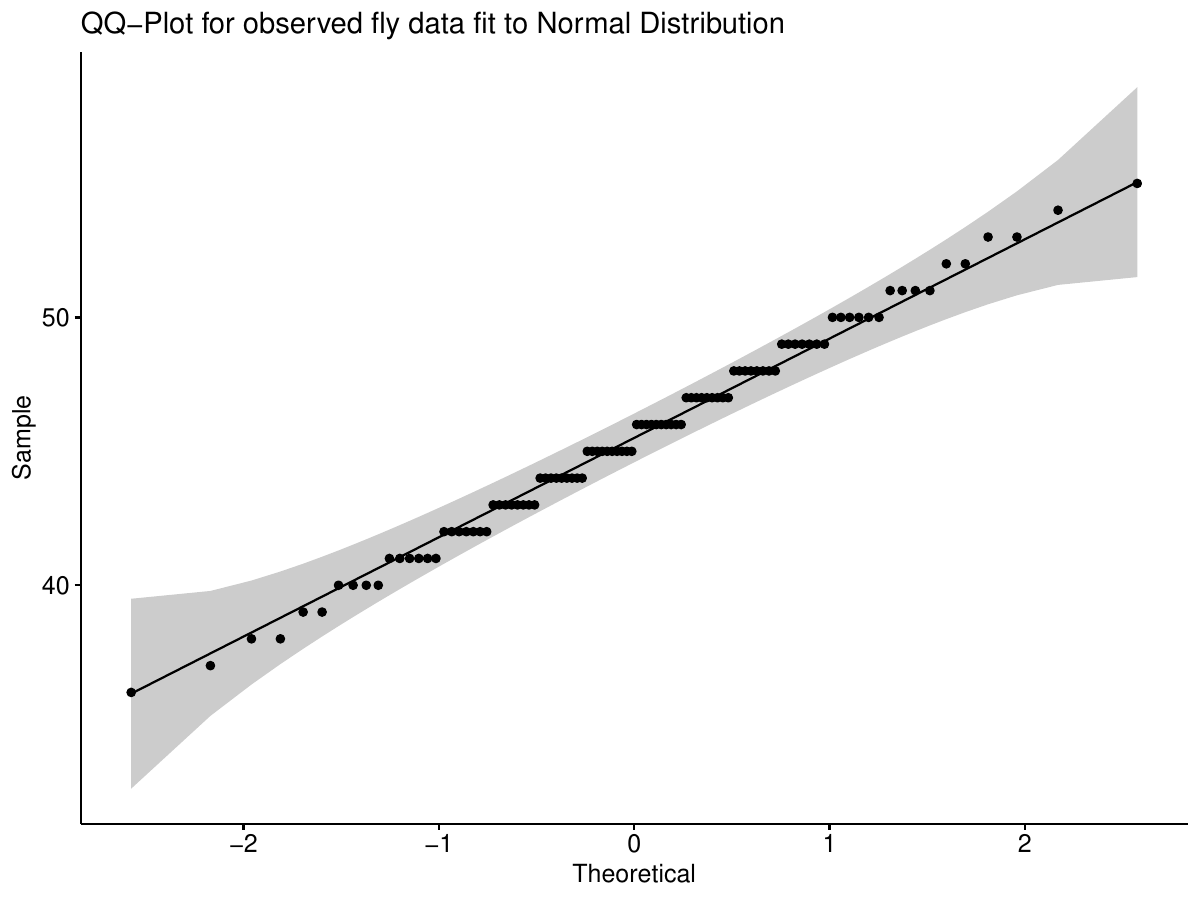}
    \caption{An example of a QQ-plot that we will further analyze in \ref{fly}. }
    \label{qqex}
\end{figure}

Quantile- Quantile plots, or "Q-Q Plots" are scatterplots of sorted sample data in ascending order, and then plot them versus quantiles calculated from a theoretical distribution. If the sample data fits the theoretical data, the scatterplot should look rather straight and align with the theoretical values. Normal Q-Q Plots are most often used in practice when fitting continuous measurement data to a distribution. This popularity reflects many statistical methodologies relying on normality. Nonetheless, Q-Q Plots can be used to analyze the fit of your sample data to any distribution \cite{QQ}. This tool is particularly useful as a first step in determining whether or not an 1D data set fits a given theoretical distribution, because it is visually simple and allows us to see where the observed  empirical quantile process deviate from the theoretical one. A Q-Q Plot coupled with a point-wise confidence band is seemingly intuitive and evident for pointing out many kind of departures away from the theoretical curve. As shown in Figure \ref{qqex} for the first illustrating example, the empirical quantile process is entirely falling within this point-wise confidence band. Is it reasonable to declare normality with confidence? The answer is negative. Since we can detect a very vivid departure pattern of being below-then-above the straight line. This fact surely reflects the limit of utility of a Q-Q plot.

As this visual appealing tool has become very conveniently and widely used, at least from data analysis perspective, it is essential to keep in mind that its utility might completely thwarted by at least two fundamental issues. First, the underlying quantile process has complex stochastic variation process along the axis of $R^1$. That is, the point-wise confidence band indeed is quite misleading sometimes. Since, even when i.i.d assumption is being assumed and based on empirical process theory \cite{hsieh95, hsieh96, hsiehTBW}, its approximation is highly dependent on its unknown distribution function and the sample size. The point-wise confidence band hardly can capture the stochasticity in data. 

Secondly, without explicitly extracting the categorical departure patterns, any data analysis result is obtained in a blind fashion. Such a result is surely subject to the risk of missing authentic information content and arriving at wrong conclusions. For instance, the aforementioned systematic departure pattern might be essential to be brought out and examined its potential impacts.

\subsection{Shapiro-Wilk normality test}
In 1965 S.S. Shapiro and M.B. Wilk introduced the Shapiro Wilk test \cite{shapiro} to determine how well a sample fits a theoretical distribution. Under the null hypothesis that a sample $x_1, ..., x_n$ came from a normally distributed population, the test statistic $W$ is constructed as:

$$W={\left(\sum _{i=1}^{n}a_{i}x_{(i)}\right)^{2} \over \sum _{i=1}^{n}(x_{i}-{\overline {x}})^{2}}$$
where $x_{(i)}$ is the ith order statistic, i.e., the ith-smallest number in the sample and
$\overline {x}=\left(x_{1}+\cdots +x_{n}\right)/n$ is the sample mean. The coefficients $a_{i}$ are given by: $ (a_{1},\dots ,a_{n})={m^{\mathsf {T}}V^{-1} \over C}$
where $C$ is a vector norm: $C=\|V^{-1}m\|=(m^{\mathsf {T}}V^{-1}V^{-1}m)^{1/2}$
and the vector $ m=(m_{1},\dots ,m_{n})^{\mathsf {T}}$ is the expected values of the order statistics of independent and identically distributed random variables sampled from the standard normal distribution; finally, $V$ is the covariance matrix of those normal order statistics. 

This testing statistic $W$ basically attempts to capture only the correlation between two vectors of observed and hypothesized order statistics. This single aspect of departure hardly accommodates other diverse aspects of potential departure existing within a large data set. 

The cutoff values for the $W$ statistic are determined by Monte Carlo Simulations. Its single p-value is likely having limited use. Since  many aspects are relevant, many p-values are needed in order to arrive at a synthesized decision.   

\subsection{Pearson's Chi-Squared ($\chi^2$) Test}
Pearson's  Chi-Squared ($\chi^2$) Test was first introduced by Karl Pearson in 1900 \cite{Pearson} and has been widely used to measure the goodness-of-fit of data to models. The chi-squared test is defined: suppose that $n$ observations being randomly sampled from a population are grouped into $K$ mutually exclusive classes with respective observed numbers $O_k$ (for $k = 1,2,…,K$), and a null hypothesis gives rise to the probability $p_k$ that an observation falls into the $k$-th class. So we have the expected numbers $E_k = np_k$ for all $k$, where $\sum_{k=1}^K p_k=1$, $\sum_{i=1}^K O_k= n\sum_{k=1}^K p_k= \sum_{k=1}^K E_k$. 

Then, given the null hypothesis is true, as $n \to \infty$, the following statistics will converge to the $\chi^2_{K-1}$ distribution with $K-1$ degree of freedom:
$$
\chi^2= \sum_{k=1}^K \frac{(O_k-E_k)^2}{E_k} = N \sum_{k=1}^K\frac{(O_k/N-p_k)^2}{p_k}.
$$

How to choose $K$ and how to form these $K$ bins are rather open issues faced in each application of this goodness-of-fit testing on any real data set. One choice was suggested in Moore \cite{Moore}: each of the bins is equal in proportion and the number of bins is often determined to be $ceiling(2∗(N^{2/5}))$. Obviously, this one-size-fits-all approach to determining bins while also creating them to all have the same proportion imposes structures on otherwise unstructured data. Using this framework for determining the test statistic completely ignores the shape information of the distribution in the data. Such an analysis likely leads us to unrealistic, if not erroneous, analysis. 

Further, this method is often used for discrete and continuous data types either in an univariate or multivariate format, making it one of the most versatile traditional methods for evaluating how well data fits a given model such as the normal distribution. However, its validity rests on several fundamental assumptions, which are likely unrealistic in many real-world applications. Its null hypothesis requires the following assumptions: a simple random sample, sufficiently large sample, independence, and adequate expected cell counts, which disqualifies many real-life data scenarios. 

Each of our examples in this paper do not adhere to the assumptions of this test as it would be impossible to assume randomness and independence when say-- measuring the length of wings of houseflies from a single experiment where the flies are subject to the same- or similar- environmental factors. Likewise, when looking at the speeds of a pitcher's pitches during a season we may not assume those conditions since from one pitch to the next the pitcher may experience fatigue or injury. Thus, using the Pearson chi-squared test is blatantly limited  from the very foundation, if not entirely inappropriate. 

\subsection{Kolmogorov - Smirnov test}
Introduced in the 1930's, the Kolmogorov-Smirnov test \cite{KS} \cite{KS2} uses the maximum distance of the empirical distribution function formed from the observed data and the theoretical distribution, to determine goodness-of-fit. The null hypothesis is that the observed data fits a given distribution. The null hypothesis also assumes that the $n$ samples are randomly drawn from the same population, samples are mutually independent, and the scale of measurement is at least ordinal. The K-S test statistic measures the largest distance between the empirical distribution function $\hat{F}_n(x)$ and the theoretical function $F_0(x)$, The test statistic is given by:

$$
D= \underset{x}{sup} |F_0(x)-\hat{F}_n(x)|
$$
The critical values of $D$ are to be found in the K-S P-value table, which is established via simulations under the null hypothesis.

This testing statistics only makes use a very limited aspect of distributional shape information of $\hat{F}_n(x)$, and leaving all other valuable departure patterns against the $F_0(x)$ unused. Further, once again, this test statistic assumes random samples and mutual independence of samples which could not apply to our examples and excludes many more real- life categorical data sets. Thus, we have reached another motivation to broaden our perspective for how such data sets can and should be analyzed: working for all data types, including categorical variable. 

The K-S test for goodness of fit has been modified many times since its introduction and in 1974 Stephens \cite{steph} provided the following equation to determine critical values and cutoff range for the test statistic given $n$ observations: 

$$
D^*= D(\sqrt{n} - 0.01 +0.85 \sqrt{n})
$$

While this cutoff range may help some in understanding what otherwise seems very "black-box" for the cutoff range, it still does not help in our case as we do not meet the base assumptions of the null hypothesis. Furthermore, in our data we have duplicate values and such data sets skew the result leading to a warning message when computing the K-S statistic in \proglang{R} reading: "Ties should not be present". This limitation once again prevents us from truly analyzing goodness-of-fit with real-world data because duplicate values are common and important to account for.

\section{CEDA method} \label{cedasection}
In this paper, we apply a new data analysis paradigm Categorical Exploratory Data Analysis (CEDA), recently developed in \cite{hsiehchou} and applied for resolving Extreme-$K$ categorical sample problem in \cite{CEDA}, to provide a methodological framework for goodness-of-fit testing. The following algorithm gives pseudo-code for how we implemented that framework for the examples in this paper. For the complete code see \cite{sabgit}. 

\begin{algorithm}[H]
\SetAlgoLined
\textbf{Input:} \textit{DATA}: 1-Dimensional observed continuous measurement data, $m$: number of simulations. \\ 
\KwResult{p-value }
\begin{enumerate}
    \item Upload observed $(n \times 1)$ data: \emph{DATA}. 
    \item  Plot a hierarchical cluster tree using the euclidean distance matrix of \emph{DATA} and method= "ward.D2". Name the tree: \emph{HC}. 
    \item  Upon \emph{HC} determine a number of $K$ clusters or branches to form a histogram such that its corresponding piece-wise linear distribution is an appropriate approximation of data's empirical distribution function, see detail developments in \cite{hsiehroy}: $K$
    \item  Upon the $K$ bins, we identify bins' endpoints by modified observed data points in order to close gaps or leaving them open as discovered, see \cite{hsiehroy} for details, and leaving the first and last bins open at the beginning and end points, respectively. Using these endpoints create a $K \times 2$ matrix: \emph{INTERVALS}. 
    \item  Simulate data $m$ times using the mean and standard deviation of \emph{DATA} and append each simulation/ mimicry to your data creating an $(m+1) \times n$ data matrix. Then count the number of data within each event that falls within the $K$ bins given by \emph{INTERVALS}, resulting in a $(m+1) \times K$ matrix: \emph{BinCount}.
    \item Take the proportion of each cell with respect to the row sum= $n$: \emph{$P_0$}
    \item Plot hierarchical clustering tree using \emph{$P_0$} on row-axis: \emph{HC2}. 
    \item Using binary nature of \emph{HC2}, evaluate odds of hitting the right branch at each internal-tree-node: Left-vs-Right, along each descending path leading to every leaf in the tree, and calculate the product of odds as a p-odds of \emph{DATA} (a leaf in our tree), that is, an odds at a internal-tree-node is the ratio of the number of leaves on the correct branch against the number of tree leaves on the incorrect branch: \emph{Podds}
    \item let our \emph{DATA} leaf number $=j$ and calculate 
    $$p-value(x_j)= \frac{\sum_{i\neq j}^{n+1} Podds(x_i)< Podds(x_j)}{n}$$
\end{enumerate}
\caption{Categorical Exploratory Data Analysis: CEDA}
\label{cedaAlg}
\end{algorithm}

We note that this algorithm is applicable to discrete and categorical data-types. In fact, any histogram reveals the categorical nature contained in data. As far as subjective choices of $K$ is concerned, the criterion as stated step 3 is rather sufficient, see the information theory detailed in \cite{hsiehroy}. There are some packages in \proglang{R} such as \pkg{NbClust} \cite{NB} that will run several tests for you and give an idea of how many clusters are likely to be appropriate. However, due to the importance of categorical nature in data, we suggest to follow the guideline discussed in \cite{hsiehroy}. One rule of thumb is that often categorical characteristics of data holds sensitive information about its distributional function. They are essential to be taken into consideration when performing goodness-of-fit testing. That is, CEDA focuses on exploring and discovering data's authentic categorical patterns, so that it allows for a much more realistic comparison point for real- world data. This is advantageous both for understanding global features of our data and for teaching others how to interpret data within suitable context.

\section{Leveraging CEDA for more visual understanding}
Here we make further use of histograms for understanding our data from the local-to-global systematic perspectives: asymmetry, humps and gaps. For visualizations through these perspectives, it is necessary to see in which bins have densities above or below our simulated data locally, and then check whether there exist local-to-global systematic patterns or not. Next we describe an algorithm of producing a heatmap that will aid in that understanding and the actual code we used in our examples can be found at \cite{sabgit}.

\begin{algorithm}[H]
\SetAlgoLined
\textbf{Input:} \textit{DATA}: observed data, $m$: number of simulations.\\
\KwResult{Heatmap}
\begin{enumerate}
    \item Apply CEDA to \textit{DATA}
    \item Make a histogram of $K$ bins on the observed data.
    \item Using \textit{BinCount} from step 5 of CEDA, subtract the $n$ simulations' bincounts from the corresponding observed data's bincounts and store it in an $(m+1) \times K$ matrix: \textit{BinDiff}\\
    \textbf{For} $i$ in $1: (m+1)\{$\\
    \setlength\parindent{24pt} \textbf{For} $j$ in $1:K\{$
    $$BinDiff[i,j] \gets \textit{BinCount}[i,j]-\textit{BinCount[1,j]}\}\}$$
    \item For each entry in \textit{BinDiff} Replace each value with a 1 if the value is positive and -1 if the value is negative. Now \textit{BinDiff} is an $(m+1) \times K$ matrix composed of \{-1,0,1\} values.  
    \item Plot a heatmap using \pkg{gplots} \cite{gplots} function \textbf{heatmap.2} with input \textit{Bindiff}. Use the parameters to highlight the observed data row.
\end{enumerate}
\caption{Using CEDA for global understanding and visualization}
\label{blocks}
\end{algorithm}

By utilizing Algorithm \ref{blocks} we can find interesting local-to-global systematic patterns in our data and explicitly pinpoint where the data deviates from the distribution we are hypothesizing. On one hand, as we will see in \ref{houseImplement}, when the heatmap does not have large blocks of a single color, we can visually see that observed densities seemingly don't against the hypothesized distribution since the simulated bin counts fall both above and below our observed data in a random manner. On the other hand, if we do see large blocks of a single color revealed in \ref{startImplement}, we know that our simulated data's bin counts consistently fall either strictly above or strictly below our observed data. Such local-to-global systematic patterns strongly indicate a global misfitting. 

In summary, CEDA emphasizes on computations that carry the capabilities of discovering potentially systematic departures by aggregating local features into global patterns. As sample size $n$ is large, such local-to-global systematic patterns of departures are keys for inferential as well as understanding purposes.

\section{Housefly Data Example} \label{fly}
\begin{figure}[H]
\centering
\begin{subfigure}{.5\textwidth}
  \centering
  \includegraphics[width=\linewidth]{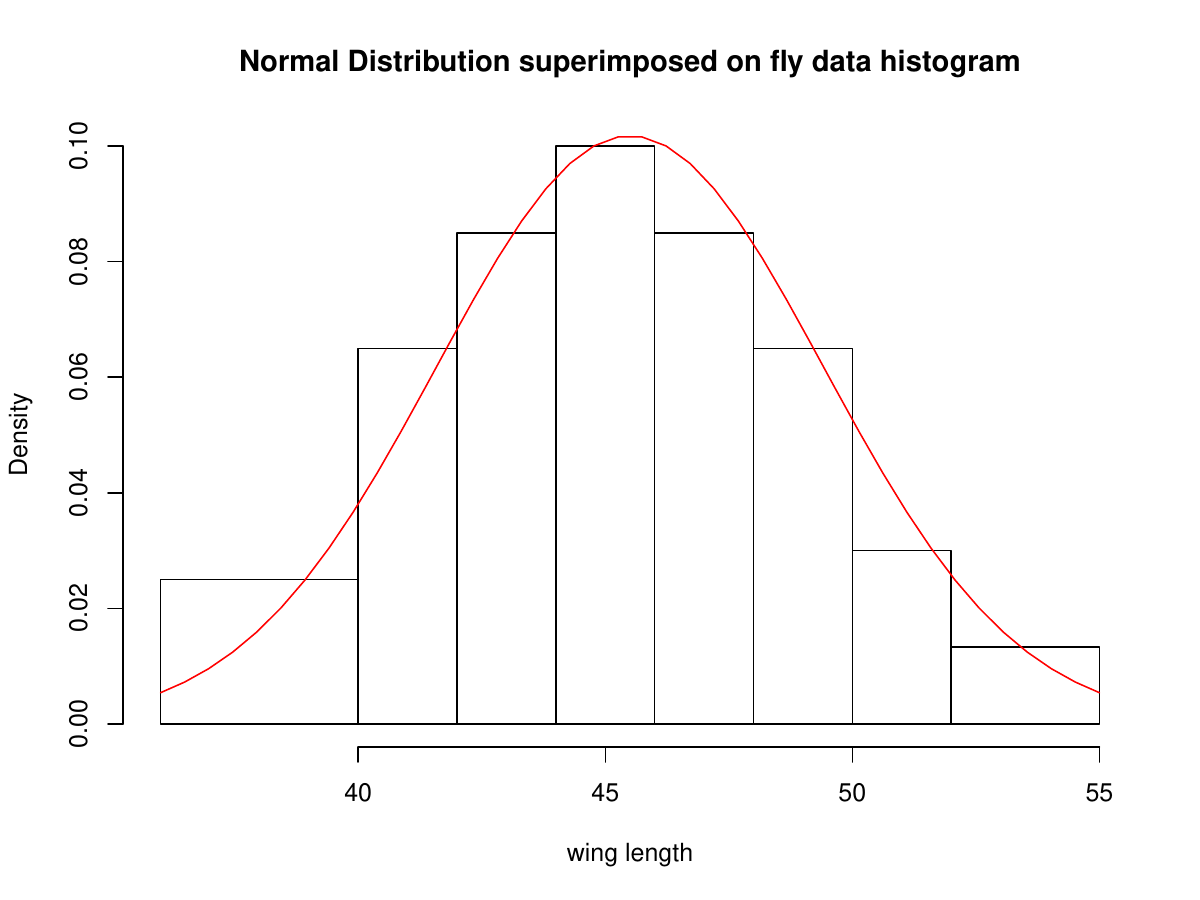}
  \caption{Histogram of fly wing length with the normal distribution superimposed in red.}
  \label{fig:density}
\end{subfigure}%
\begin{subfigure}{.5\textwidth}
  \centering
  \includegraphics[width=\linewidth]{figures/flyQQ.pdf}
  \caption{QQPlot comparing observed data's distribution to theoretical normal distribution}
  \label{fig:qq}
\end{subfigure}
\caption{Traditional methods demonstrating \emph{normality} visually}
\label{fig:tradfly}
\end{figure}

 We begin with a data set distributed by 'Quantitative Environmental Learning Project' \footnote{\url{https://seattlecentral.edu/qelp/sets/057/057.html}} named Data Set \#057 \cite{flydata}. This data set contains 100 observations of housefly wing lengths ($\times$.1mm) and yields a nearly perfect normal distribution with mean $\mu = 45.5$ and standard deviation $\sigma= 3.919647$. Using this data we show a density plot in Fig. \ref{fig:density} and Q-Q plot in Fig. \ref{fig:qq} to demonstrate visually that this data is, in fact, looking like normal distributed.

Further, the Shapiro-Wilk Test returns $$ p-value= 0.8761.$$ Since it is well over 0.05 threshold, we don't object that the data fits a normal distribution. 

However, as previously discussed at the end of Section \ref{cedasection}, when taking in categorical nature of real-world data into considerations, it might improve our understanding if instead of holding the data to the standard of theoretical, continuous distributions, we compare this data with a random sample generated from the normal distribution with parameters given by the original data's statistics.


Note, the histogram in Fig. \ref{fig:density} does not have equal intervals and that is because we used the bin intervals given by the CEDA algorithm to plot the observed data. 

Using the Pearson Chi-Square test on the data with $ceiling(2 * (100^{2/5})$ classes, we get: 
$$P = 11.8,\text{ } p-value = 0.2987$$

Using the Kolmogorov - Smirnov test we get:

$$D = 0.050752,\text{ } p-value = 0.9589$$

However, we also get a warning message that ties should not be present for this test because there are repeated values in our observed data. 

Here, this data is by no stretch of the imagination "random" since it was the measurements of housefly wings in a closed experiment. The notion of randomness in sampling is simply not feasible in this setting because the environment is controlled and the specimens are all accounted for in the data acquisition. Moreover, the observations might not be independent or identically distributed due to this data coming from a single experiment. Thus, none of the tests we just did should have been perfectly certain if we were to adhere to the assumptions of each test's null hypothesis. To alleviate the unattainable assumptions a little bit from our analysis and thereby ensure we do not violate any, we reevaluate this data using CEDA paradigm.

\subsection{Implementing CEDA Algorithm \ref{cedaAlg}}

\begin{figure}[H]
    \centering
    \includegraphics[width= .6\textwidth]{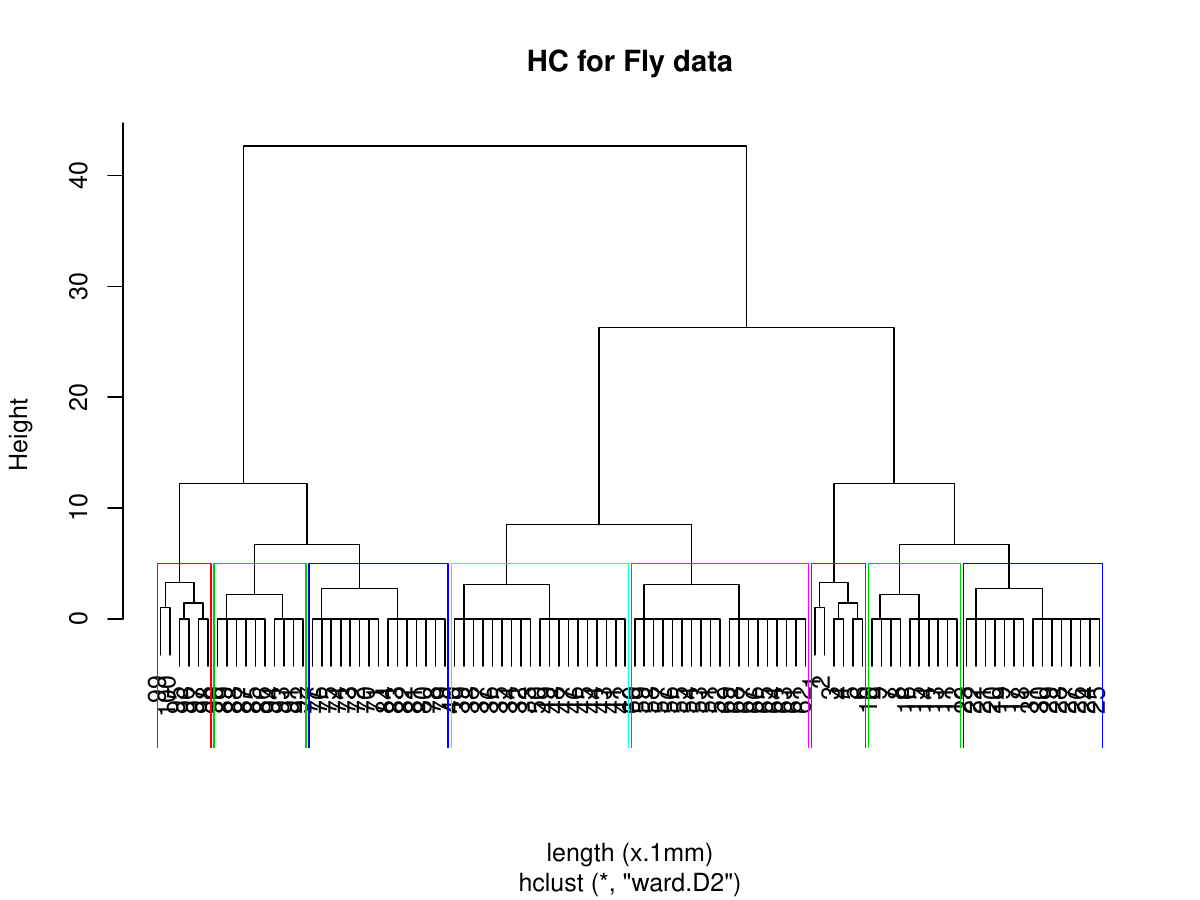}
    \caption{HC tree for the observed data cut into 8 clusters.}
    \label{fig:fly8bins}
\end{figure}

Plotting the hierarchical clustering tree corresponding to our observed data we determine that $K=8$ clusters is appropriate. Then we cut our tree into 8 clusters as shown in Figure \ref{fig:fly8bins} and continue implementing the algorithm. 

\begin{figure}[H]
    \centering
    \includegraphics[width= .8\textwidth]{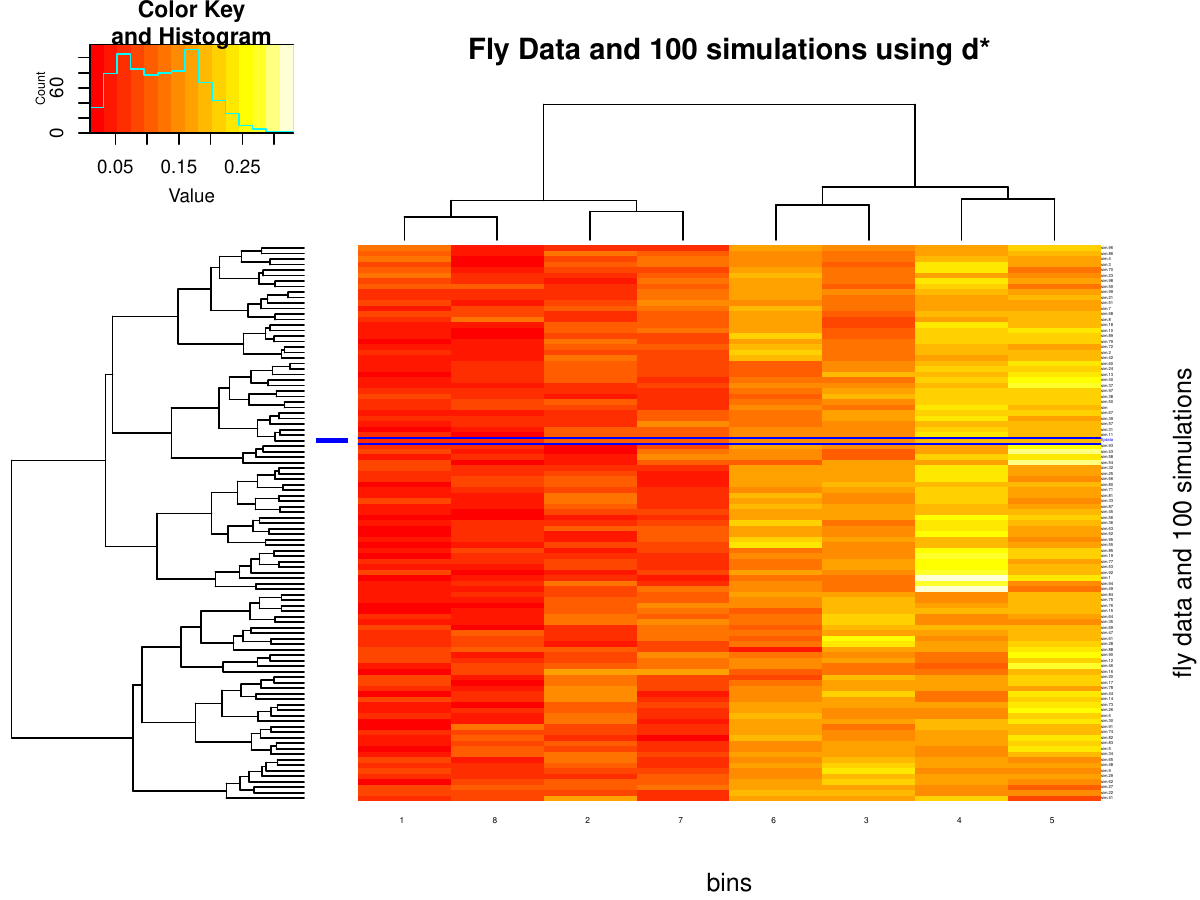}
    \caption{HC tree for $P_0$ separated into 8 clusters and our observed data highlighted in blue.}
    \label{fig:fly100}
\end{figure}

Now, using CEDA we simulate the data 100 times and plot it's subsequent \emph{$P_0$} hierarchical clustering tree. For this example we will show the hierarchical cluster tree on the vertical axis of the heatmap produced by \emph{$P_0$}. This tree represents how each of the rows in \emph{$P_0$} are clustered using method= 'ward.D2'. To do this we use \pkg{gplots} \cite{gplots} function \textbf{heatmap.2} with input \emph{$P_0$} and show the result on Figure \ref{fig:fly100} with our observed data highlighted in blue. 

Using CEDA and the HC tree shown in Figure \ref{fig:fly100}, we find that the p-value for this observed data is $$
p-value= 0.73
$$
 once again affirming that this observed data likely fits a normal distribution. 
 
 \subsection{Implementing Algorithm \ref{blocks}}\label{houseImplement}
 
\begin{figure}[H]
    \centering
    \includegraphics[width= .8\textwidth]{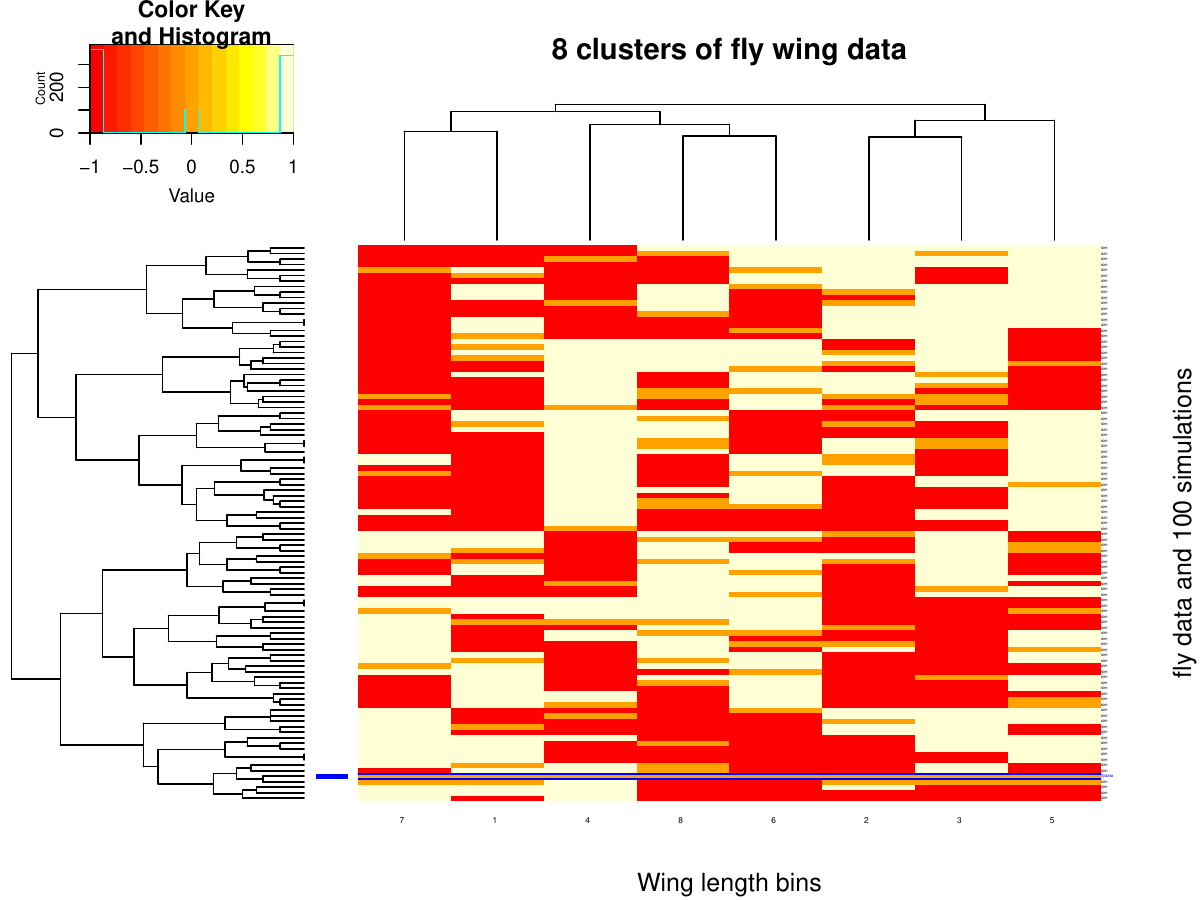}
    \caption{HC tree for \textit{BinDiff} separated into 8 clusters and our observed data highlighted in blue.}
    \label{fig:fly100proto}
\end{figure}

 Now we look at the heatmap produced by algorithm \ref{blocks}. As mentioned at the introduction of this algorithm \ref{blocks}, if we see that the 0, 1, -1 values represented by red, white, and orange respectively are mixed within each bin and the heatmap does not show large blocks of a single color, we may believe our observed data fits the distribution. In Figure \ref{fig:fly100proto} we observe that our data blends very well with the 100 simulations, further supporting our conclusion of normality. Since this heatmap is extremely varied and no bins are overwhelmed by a single color, we can quickly deduce that our data suits the normal distribution well. 
 
 This variation on CEDA gives a more global perspective on how this data fits within the simulations and in a sense mirrors the Q-Q plot. While the heatmap generated using the CEDA $P_0$ matrix gives more detailed differences between individual datum, Figure \ref{fig:fly100proto} allows us to quickly and accurately understand how our data fits with respect to the simulation set bin by bin. For example, using Figure \ref{fig:fly100proto}'s first column we can see that roughly the top half of the data represented in bin 7 falls below or at the bin count of the observed data shown with a block of red and orange. Furthermore, in bin 7 we see that this observed data shares many branches in the dendrogram with simulations' whose bin counts are larger than the observed shown by the large presence of yellow. Returning to the Q-Q plot in Figure \ref{fig:tradfly} we notice that on the far right end of the plot, where bin 7 and 8 are located, our data is above the theoretical and it makes intuitive sense that our data is surrounded by yellow in bin 7. However, bin 8 shows this data surrounded by red and orange, due to the observed data's eighth bin having few data points, as illustrated in the last bar of the histogram in Figure \ref{fig:tradfly}. Similar observations can be made of bins 1 and 2 in Figure \ref{fig:fly100proto} as our traditional methods shown in Figure \ref{fig:tradfly} indicate that the observed data falls below the theoretical in the first couple bins. 


\section{Starting Speed Pitching Example}
\label{pitch}

\begin{figure}[H]
\centering
\begin{subfigure}{.5\textwidth}
  \centering
  \includegraphics[width=\linewidth]{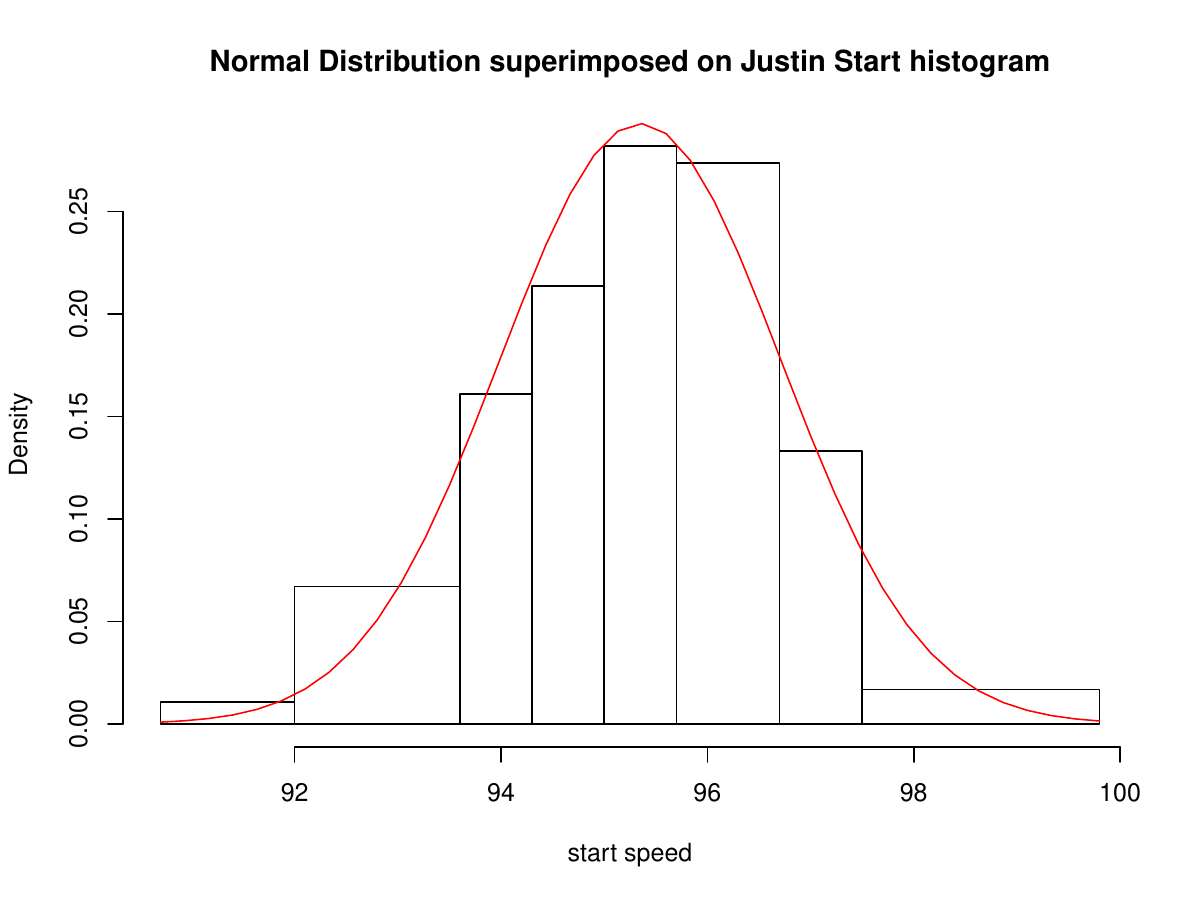}
  \caption{Histogram of start speed data with the normal distribution superimposed in red.}
  \label{fig:densitystart}
\end{subfigure}%
\begin{subfigure}{.5\textwidth}
  \centering
  \includegraphics[width=\linewidth]{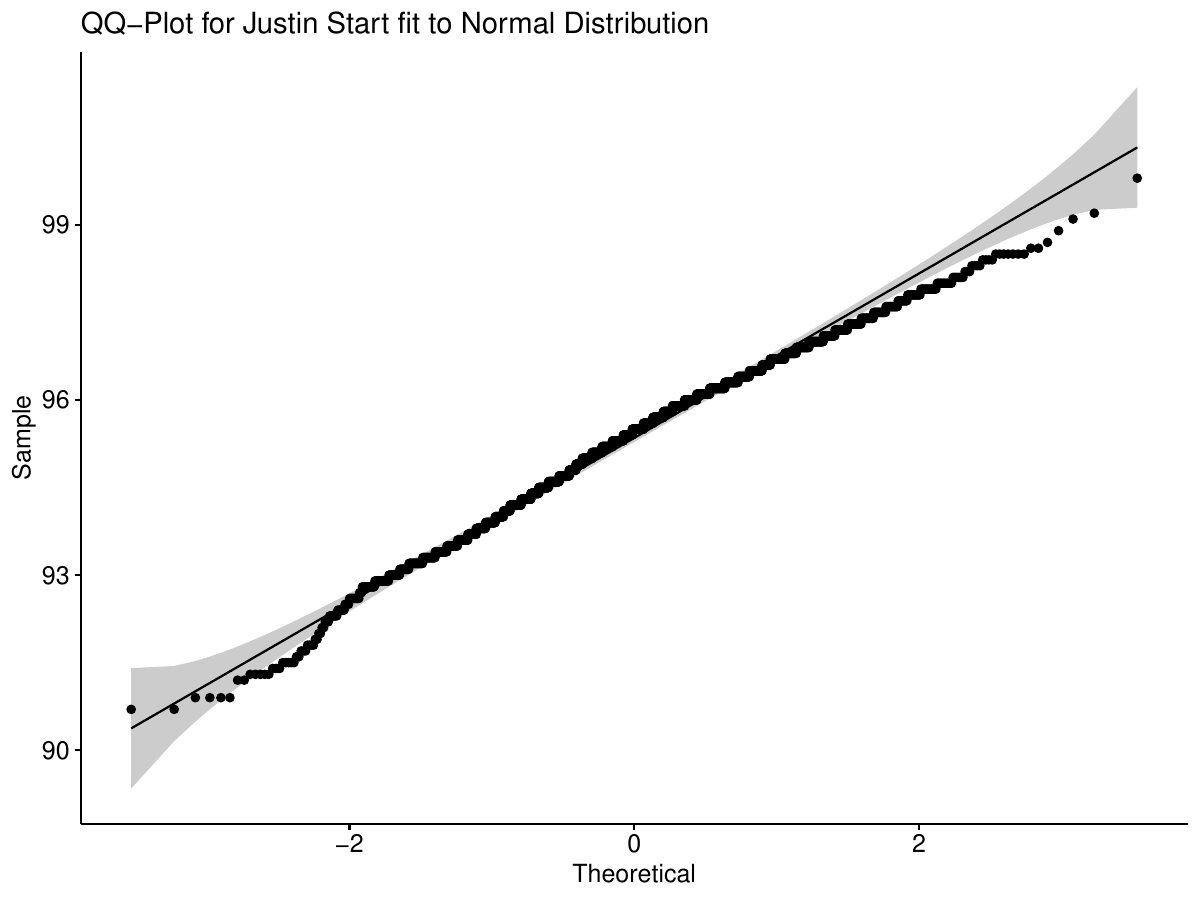}
  \caption{QQPlot comparing observed data's distribution to theoretical normal distribution}
  \label{fig:qqstart}
\end{subfigure}
\caption{Traditional methods demonstrating \emph{normality} visually}
\label{fig: tradstart}
\end{figure}

Now we shift our attention to a larger data set collected from the 2017 Major League Baseball season. In this example we consider the starting speed of every pitch the MLB pitcher Justin Verlander threw during all official 2017 games. This data is significantly larger than the previous example, because here we have 2432 observations of Verlander's starting pitch speed. It is reasonable to assume that over the course of nearly 2.5 thousand pitches there was a decent amount of variation and perhaps this data fits a normal distribution. As in the last example we will begin by observing traditional methods and plot a histogram with the normal distribution corresponding to this data superimposed onto it. Furthermore, we plot a Q-Q plot of the start speed data and display both in Figure \ref{fig: tradstart}. 

Again, note that the histogram intervals are not uniform as they match the 8 bin intervals we use when implementing CEDA.

Using this data we find the mean $\mu = 95.34992$ mph and standard deviation $\sigma =1.361738$ mph. By observing these plots we notice immediately that the histogram is skewed to the left and the data at the tails are below the theoretical expectation. Naturally, someone with practice determining goodness of fit might conclude that it is unlikely to fit the distribution. For extra certainty we compute the Shapiro -Wilk p-value and discover: 
$$p-value = 1.863691e-10$$.

The p-value here is significantly lower than .05 and allows us to confirm our visual observations: this data is not normally distributed. 

Using the Pearson Chi-Squared test on the data with $ceiling(2 * (2432^{2/5}))`$ classes, we get: 
$$P = 401.77,\text{ } p-value < 2.2e-16$$

Using the Kolmogorov - Smirnov test we get:

$$D = 0.050792, \text{ }p-value = 7.102e-06$$

However, we also get a warning message that ties should not be present for this test because there are repeated values in our observed data. 

Once again, we note that none of these methods were appropriate for this data set since this real-world data does not satisfy the random and I.I.D. conditions assumptions. The data is from one player during a single year of playing and therefore cannot be random. For instance, many data points might come from a single game and it is reasonable to assume that over the course of performing the player might get fatigued, strained, or injured to some degree. Thus, looking at this data at only the local level instead of on a global test scale we lose information because we are only considering a single type of departure from the theoretical. Here, it is not very reasonable to subject this data to a strictly theoretical point of comparison, but instead other data sets that are generated by the normal distribution using the parameters given by our observed set. Rather than violating the base assumptions of these tests let's move forward with evaluating goodness-of-fit using the CEDA framework and methodology.

\subsection{Applying CEDA Algorithm \ref{cedaAlg} to Start Speed Data}

\begin{figure}[H]
    \centering
    \includegraphics[width= .7\textwidth]{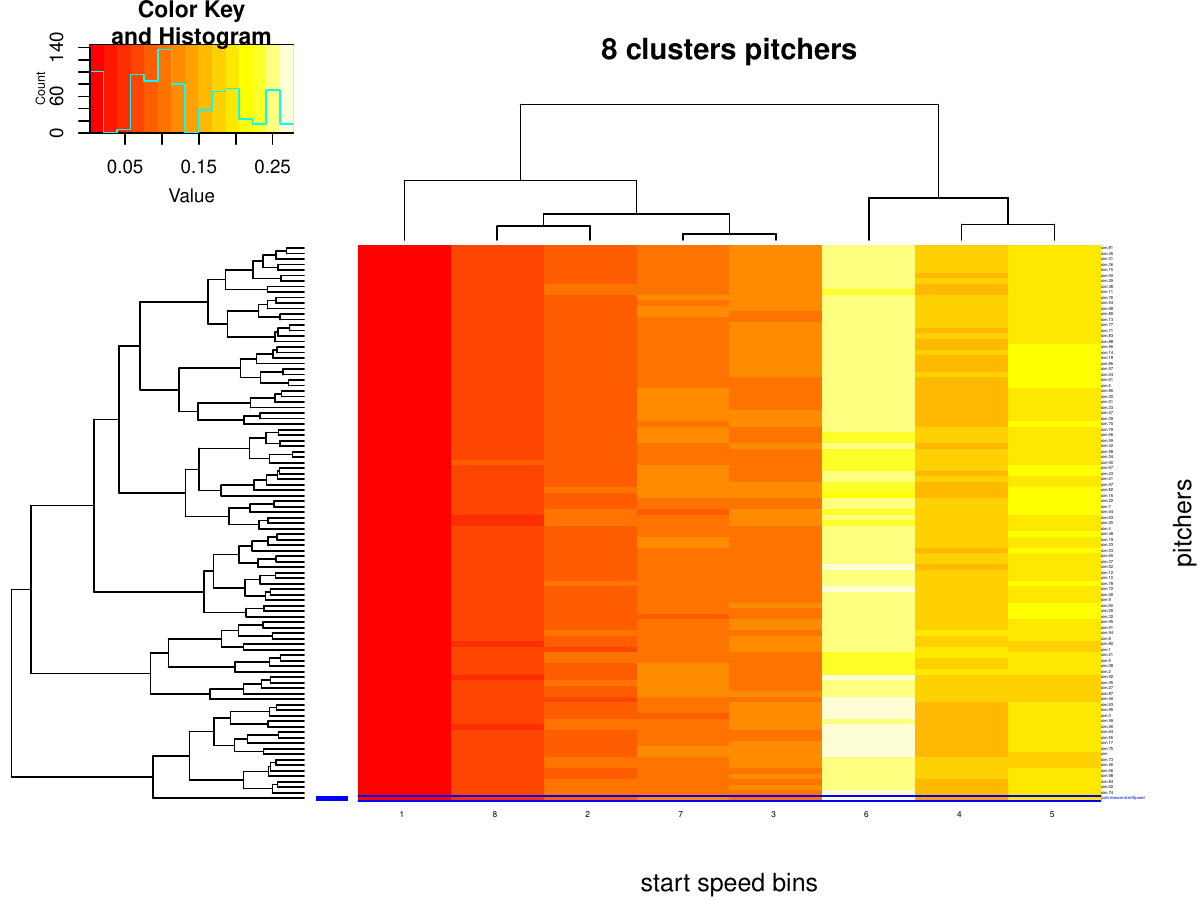}
    \caption{HC tree for $P_0$ separated into 8 clusters and our observed data highlighted in blue.}
    \label{fig:heatstart}
\end{figure}

Just as in the previous example we created a hierarchical clustering tree on our observed data and decided to cut it into $K=8$ clusters. We omitted that figure simply because it can be found on the vertical axis of Figure \ref{fig:heatstart}. It is only a coincidence that 8 clusters was appropriate in both this example and the previous. Here we could also choose $K=5$ or $K=14$, but 5 seemed too general and 14 too narrow for the clusters. 

Next we continue with CEDA and choose to simulate Verlander's 2432 pitches 100 times. Computing the $P_0$ matrix over the 100 simulated pitchers and Verlander versus the 8 bins we sorted them into we are able to produce Figure \ref{fig:heatstart}. 

Finishing the algorithm we compute:
$$p-value = 0.$$
This result aligns with the traditional methods and again allows us to confirm that our observed data does not fit a normal distribution. Moreover, observing the HC tree of the pitchers we see that our observed data branches off into a leaf very early in the branch splitting confirming it is not similar to the simulated data. Now we have a p-value that was computed without violating any necessary assumptions and can conclude a bad fit with sincere confidence. 

\subsection{Applying Algorithm \ref{blocks} to Start Speed Data}\label{startImplement}

\begin{figure}[H]
    \centering
    \includegraphics[width= .8\textwidth]{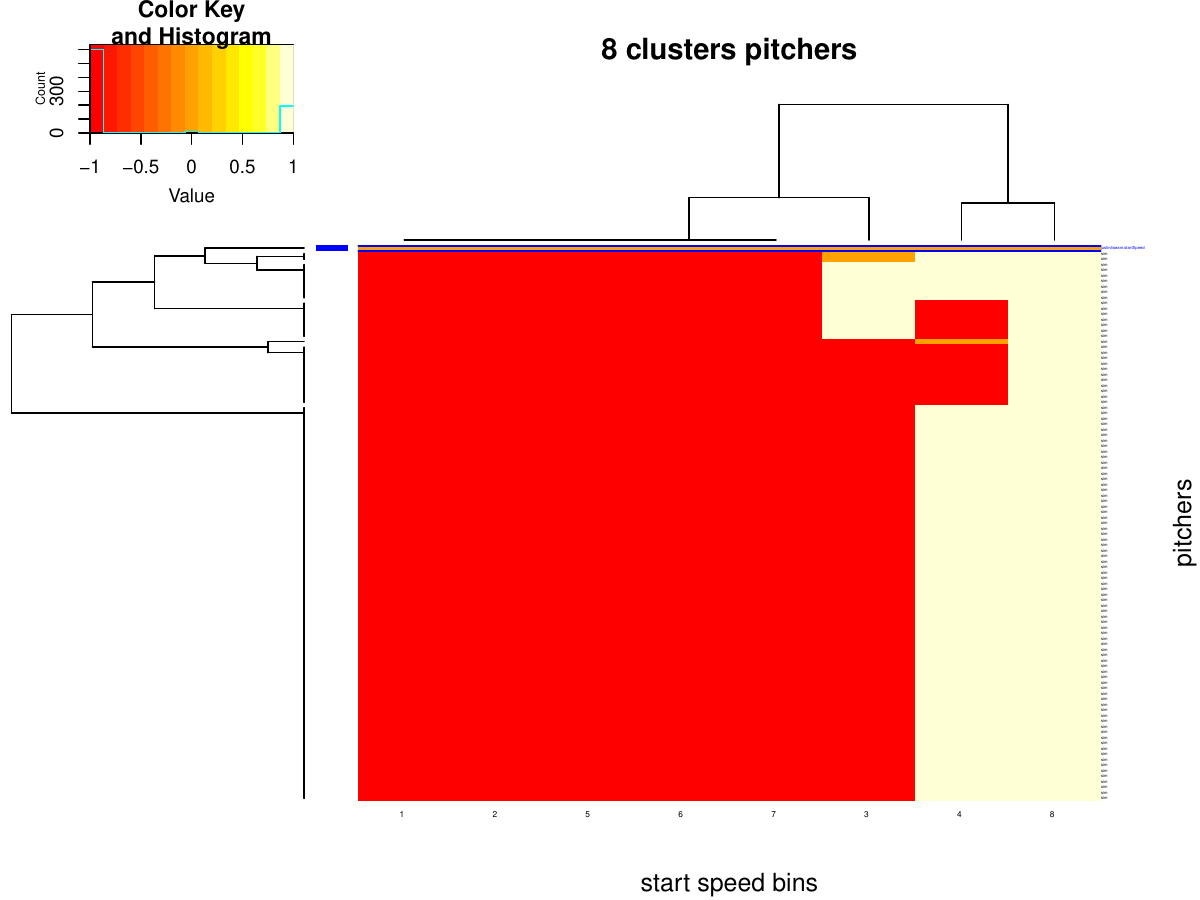}
    \caption{HC tree for \textit{BinDiff} separated into 8 clusters and our observed data highlighted in blue.}
    \label{fig:protostart}
\end{figure}

As in the housefly example, we wish to understand our data in the context of simulations derived from it at a glance. Thus, we apply algorithm \ref{blocks} to our start speed data and attempt to glean some insights. The algorithm yields Figure \ref{fig:protostart} and we immediately see that our data does not blend with the rest. Here the bin by bin comparison is dramatic.

In bins 1,2,5,6 and 7 all we see is red showing that every simulation bin count is less than the observed bin count. Bins 3 and 4 offer a small amount of diversity and bin 8 shows every single simulation had higher bin count than our observed. The uniformity in each of the columns visually emphasizes how out of place our data is within the normal distribution. Again, the heatmap corroborates the information we could deduce from the traditional methods for visually inspecting our data, but offers significantly more insight for the global shape and distribution of our data. This method of analysis does not flatten our data as in the traditional methods because we use high dimensional differences via $P_0$ when producing the heirarchical clustering tree needed to produce the bin counts represented by this heatmap.

We are also able to see the type of trend we saw with the Q-Q plot with the observed data falling below the theoretical near the tails, but now we have greater understanding of how our data falls below in the space of randomly generated normal data. Instead of the limited local perspective given by the Q-Q plot we see that our data falls below at the tails for every one of our 100 simulations. We also see that the far right of our data is rather sparse because bin 8 has more variety due to the sparsity of the data and therefore low bin count.

\section{Ending Speed Pitching Example}
\label{pitchEnd}
We wrap up our examples with more data from the 2017 MLB season. Here we consider the ending speed recordings of Justin Verlander's pitches during the season. In correspondence with our last example, there are 2432 end speed observations. We begin once again by partaking in traditional methods and generate a histogram with the natural distribution superimposed and a Q-Q plot of our End speed data against the theoretical normal distribution.

\begin{figure}[H]
\centering
\begin{subfigure}{.5\textwidth}
  \centering
  \includegraphics[width=\linewidth]{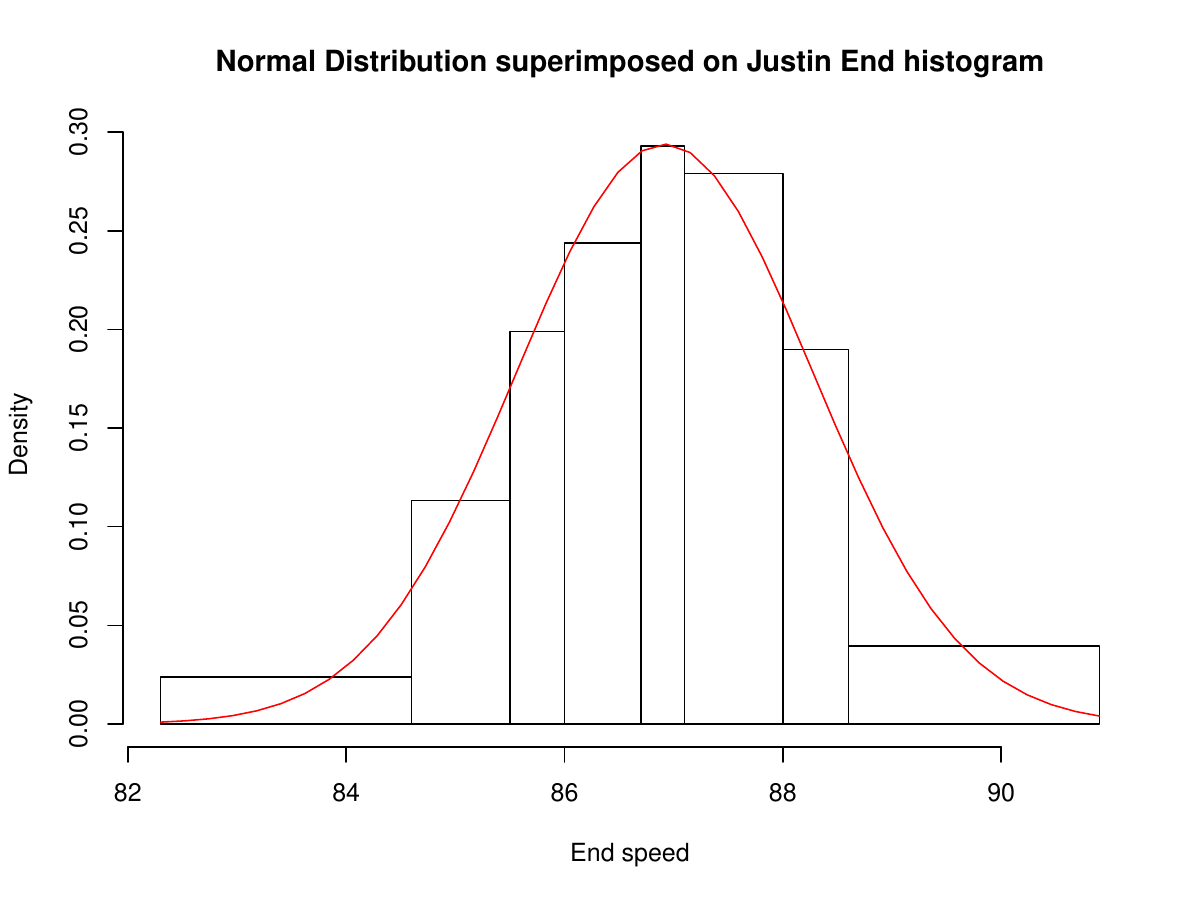}
  \caption{Histogram of end speed data with the normal distribution superimposed in red.}
  \label{fig:densityend}
\end{subfigure}%
\begin{subfigure}{.5\textwidth}
  \centering
  \includegraphics[width=\linewidth]{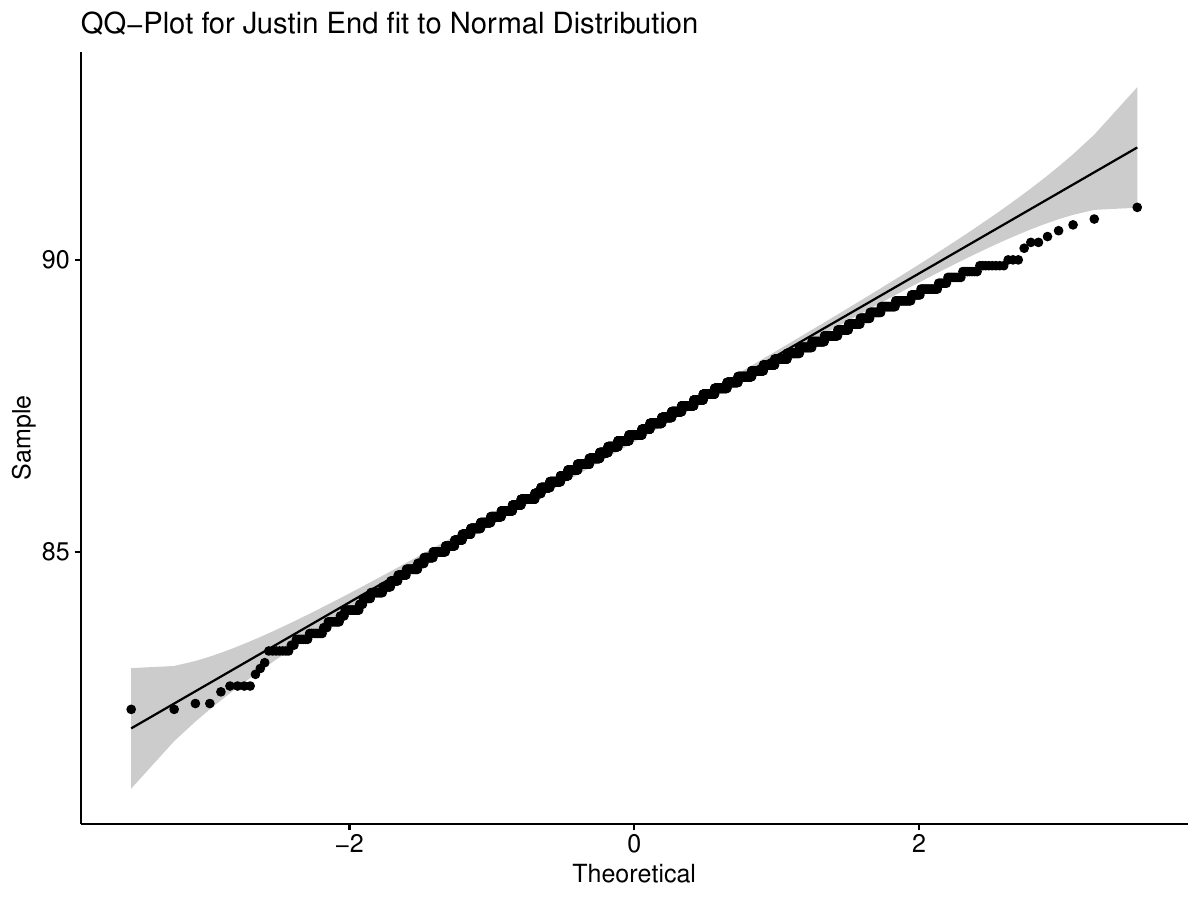}
  \caption{QQPlot comparing observed data's distribution to theoretical normal distribution}
  \label{fig:qqend}
\end{subfigure}
\caption{Traditional methods demonstrating \emph{normality} visually}
\end{figure}

As before, the histogram intervals are not uniform because they reflect the intervals found using CEDA. 

Notice, the histogram is not so obviously skewed this time, but still leans toward being skewed to the left. Furthermore, the Q-Q plot looks strikingly similar to figure \ref{fig:qqstart}. Of course we would expect consistency between the start and end speed of a pitch given the laws of physics, but the $K=8$ intervals represented by each bar in the histogram are spread much differently than they were in Figure \ref{fig:densitystart}. So, it is worth continuing and learning how this data set compares to the normal distribution. Next, we compute the Shapiro Wilk test:
$$p-value = 3.524864e-08 $$

Using the Pearson Chi-Square test on the data with $ceiling(2 * (2432^{2/5})$ classes, we get: 
$$P = 466.72,\text{ } p-value < 2.2e-16$$

Using the Kolmogorov - Smirnov test we get:

$$D = 0.040323, \text{ }p-value = 0.0007353$$

As in the previous example we find that our data does not fit the normal distribution since the p-value is much less than .05. Again, we reiterate that our data does not satisfy the base assumptions for these tests. So, in reality we should not be using them, but in an effort to compare the commonly used tools in statistics, we provide them here. However, in an effort to better understand how this data varies from the start speed data analysis we forge ahead and implement CEDA.

\subsection{Applying CEDA Algorithm \ref{cedaAlg} to End Speed Data}

\begin{figure}[H]
    \centering
    \includegraphics[width= .6\textwidth]{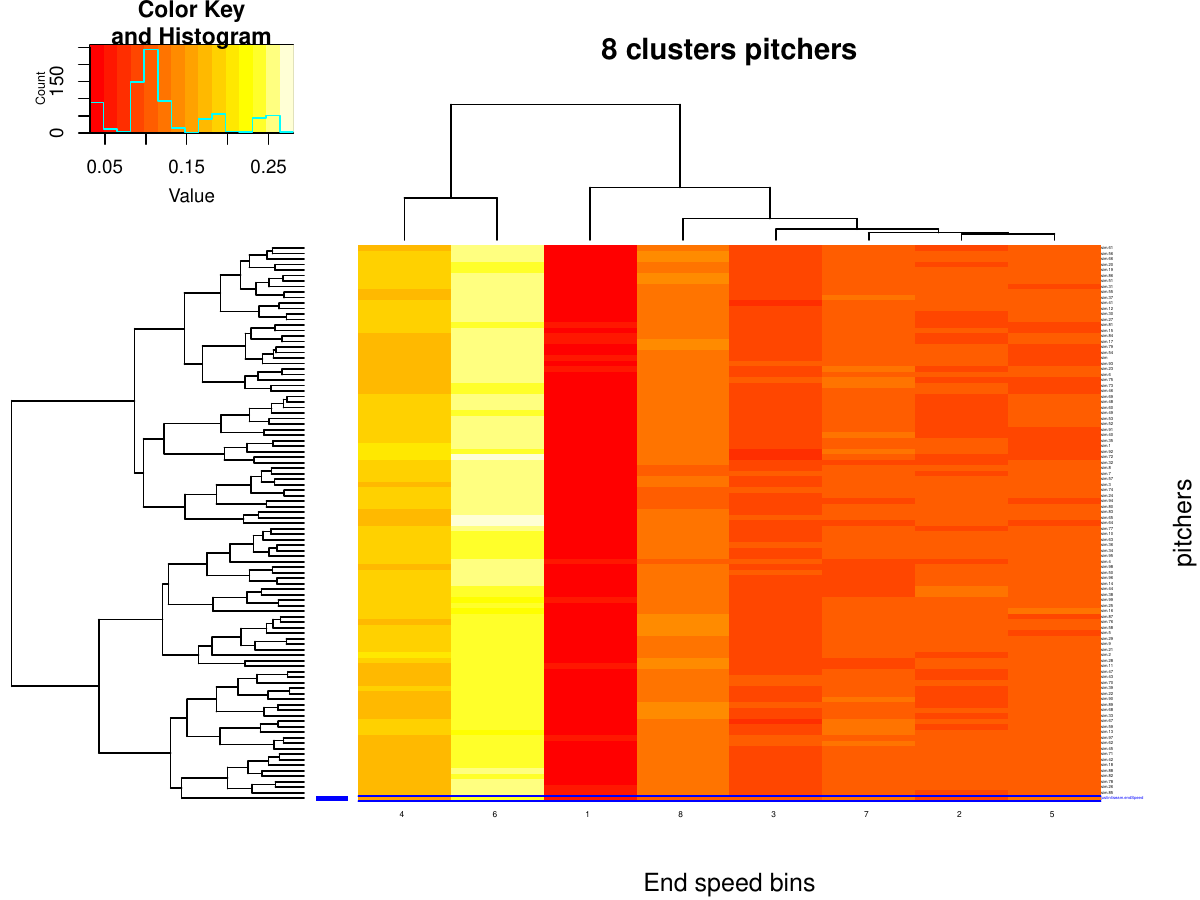}
    \caption{HC tree for $P_0$ separated into 8 clusters and our observed data highlighted in blue.}
    \label{fig:heatend}
\end{figure}

Without rehashing everything from the previous two examples we dive in and produce Figure \ref{fig:heatend} using $K=8$ clusters for consistency with the last example, producing a CEDA heatmap for Justin Verlander's End Speed data and 100 random normal simulations.

Using CEDA we find:

$$p-value = 0$$
and confirm again that our observed data is not normally distributed. This conclusion is further supported by the dendrogram showing the observed data splitting very early and therefore being unlike any simulation. That early split drives the CEDA Podds value down quickly for the node corresponding to our observed data.

\subsection{Applying Algorithm \ref{blocks} to End Speed Data}

\begin{figure}[H]
    \centering
    \includegraphics[width= .6\textwidth]{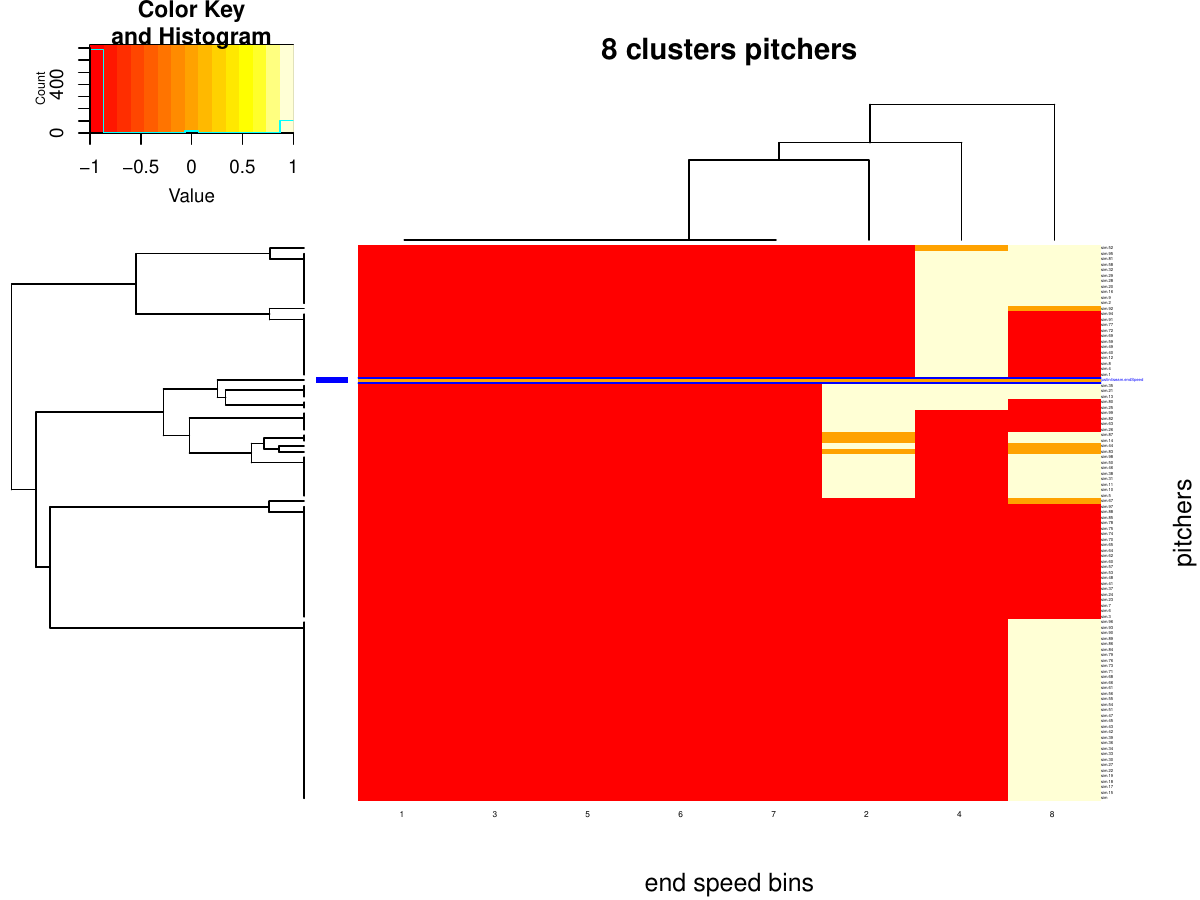}
    \caption{HC tree for \textit{BinDiff} separated into 8 clusters and our observed data highlighted in blue.}
    \label{fig:protoend}
\end{figure}

Now we use algorithm \ref{blocks} to show for a final time in this paper that upon first glance, Figure \ref{fig:protoend} demonstrates that our data is not normally distributed. 

This heatmap is shows more variation in bins 2, 4, and 8 than we saw in Figure \ref{fig:protostart}, reinforcing our point: this visualization conveys more about bin by bin subtleties than the traditional methods and allows us to better analyze our 1-D categorical data. Here we see no bin column dominated by white unlike in the previous example. That shows there are not any bins where every simulation has a higher bin count than our observed bin count. This information is useful for understanding how this data varies from the start speed data and allows us more insight into where the observed data lives in the space of random simulations.  
\section{Discussion regarding size of data}
In our three examples we had observed data sets of sizes 100 and 2432. One of the many advantages of following the CEDA framework is that large data is handled exceptionally well.Using traditional graphical techniques for evaluating goodness-of-fit such as the Q-Q plots in our examples loses usefulness when our data set is large as in examples \ref{fig:qqstart} and \ref{fig:qqend}. The plots are hardly distinguishable in these examples simply due to overcrowding. As we hope to learn more about where the data departs from the theoretical, the CEDA figures provide significantly more insight about the subtle patterns. Furthermore, with large data, mimicry can contextualize our observed data in the space of normally distributed events, while traditional methods only show how it compares to the continuous theoretical model. Thus, the traditional models give a localized perspective of goodness-of-fit while the CEDA machine learning model gives a much more global perspective of goodness-of-fit.

However, when data is not large, such as in the housefly example, each cluster is rather small, particularly near the tails of our distribution. This creates more noise as we simulate the data because it becomes more likely to have empty clusters or "bins" in our histogram and we could end up with gapped histograms. In the hierarchical clustering tree those simulations with gapped histograms will appear as significantly different from the rest therefore creating more branches and noise. As we increase data size we decrease noise by having higher bin counts per bin.  

In today's data-driven society, there is no shortage of \emph{BIG} data. Using CEDA and derivative methods, such as algorithm 2, gives us important global perspectives on categorical 1-D data while simultaneously making dauntingly large data more manageable.

\section{Conclusion}
Each of the traditional methods we have inspected in this paper have assumptions that our data and most real-world 1-dimensional categorical data cannot satisfy. Given the nature of our 1-dimensional categorical data, random and I.I.D. assumptions are impossible and therefore traditional methods exclude real-world data often. Not only are these methods not truly applicable to our data if we are to follow the necessary assumptions, but they also limit our understanding of the data to a narrow viewpoint. Many of these methods such as the K-S test and the Pearson Chi-squared test are limited to one type of deviation of the Empirical Distribution Function (EDF). In the setting we have considered we have unstructured data and the analysis should be high dimensional to account for any potential trends with our observed data. There are many potential structured deviations that can occur and our approch is more systematically geared toward finding hidden patters. Through our construction we not only adhere to the rules by keeping no assumptions on our data but we also allow space for discovery. The traditional methods impose innapropriate structure on our data and in doing so we lose the story the data may be trying to tell us. For instance, the Pearson chi-squared bins are all evenly distributed and such a one-size fits all approach does not take into account the data's distribution at all. With this CEDA method we are able to see how the data clusters and therefore how our bin intervals should be defined. Instead of imposing such structure this method allows the data to lead us to the correct bin intervals rather than steering it to an ill-fitting model.

 Using the CEDA method to evaluate normality of a sample gives us more information about where the sample deviates from the theoretical normal distribution and allows us to learn how likely our sample is in a set of randomly generated normal distributions. We notice that as we increase our data size, from 100 observations in the housefly example to 2432 observations in the pitching examples, we are able to decrease noisiness in our models. This noise reduction makes intuitive sense, because as our number of observations increases, each of our histogram bin counts increases and allows for a potentially closer fit or as we saw in the pitching example, makes the lack of fit more apparent. Thus, in today's data-driven world, using CEDA presents many benefits in understanding where a distribution model may fit a sample and where it does not. Our ability to see where our sample lands in the space of randomly generated sample with the intended distribution gives a visual understanding of how well our data fits a given distribution. 
 
 This model is applicable to any distribution and therefore can be a useful tool for teaching distributions in statistics and applied math. While the traditional methods yield p-values with respect to the theoretical distribution, our CEDA p-value sheds light on how likely a sample is to occur within some number of random samples generated with our theoretical distribution and our heatmaps show us visually how our sample fits within the space. These observations promote general understanding of likelihood, data mimicry, and are largely applicable to real-world data scenarios. Using this method, students could compare different samples within a set and their p-values will correspond to their likelihood within the given data set rather than a potentially unattainable theoretical model. If we are to transform our methods for teaching data analysis to reflect real-world scenarios and prepare students to work with big data, the CEDA method seems more appropriate. Furthermore, utilizing the algorithm gives users a more hands-on experience with the data while the heirarchical clusters give visual and mathematical justification for the p-values. That justification is not matched by the traditional black-box methods of determining p-values taught to most students.

\section{Acknowledgements}
This material is based upon work supported by the National Science Foundation Graduate Research Fellowship under Grant No. 1650042.

\bibliographystyle{alpha}

\bibliographystyle{unsrt}

\end{document}